\pdfoutput=1

\documentclass[11pt]{article}

\usepackage{EMNLP2022}

\usepackage{times}
\usepackage{latexsym}

\usepackage[T1]{fontenc}

\usepackage[utf8]{inputenc}

\usepackage{microtype}

\usepackage{inconsolata}

\usepackage{tabularx}
\usepackage{times}
\usepackage{latexsym}
\usepackage{booktabs}
\usepackage{xcolor, soul}
\usepackage{graphicx}
\usepackage{amsmath}
\usepackage{subcaption}
\usepackage{amssymb}
\usepackage{pifont}
\definecolor{OliveGreen}{rgb}{.8,1,.8}
\sethlcolor{OliveGreen}

\title{NaturalAdversaries: \\ Can Naturalistic Adversaries Be as Effective as Artificial Adversaries?}

\newcommand\ai{$^\diamondsuit$}
\newcommand\msr{$^\heartsuit$}
\newcommand\uw{$^\spadesuit$}

\author{Saadia Gabriel\uw \space\space\space Hamid Palangi\msr \space\space\space Yejin Choi\uw\ai  \\
\uw Paul G. Allen School of Computer Science \& Engineering, University of Washington \\
\msr Microsoft Research \\
\ai Allen Institute for Artificial Intelligence \\ 
\texttt{\{skgabrie, yejin\}@cs.washington.edu} , \texttt{hpalangi@microsoft.com} \\
}

\begin{document}

\maketitle

\begin{abstract}
While a substantial body of prior work has explored adversarial example generation for natural language understanding tasks, these examples are often unrealistic and diverge from the real-world data distributions. 
In this work, we introduce a two-stage adversarial example generation framework (NaturalAdversaries), for designing adversaries that are effective at fooling a given classifier and demonstrate natural-looking failure cases that could plausibly occur during in-the-wild deployment of the models. 

At the first stage a token attribution method is used to summarize a given classifier's behaviour as a function of the key tokens in the input. In the second stage a generative model is conditioned on the key tokens from the first stage. NaturalAdversaries is adaptable to both black-box and white-box adversarial attacks based on the level of access 
to the model parameters. Our results indicate these adversaries generalize across domains, and offer insights for future research on improving robustness of neural text classification models. 
\end{abstract}

\section{Introduction}



Transformer models have gained prominence in NLP research due to their powerful performances on leaderboards. However, numerous studies have shown these neural models are brittle, frequently taking shortcuts to reach decisions rather than reasoning about the underlying semantics correctly  \cite{Geirhos2020ShortcutLI,LeBras2020AdversarialFO} or failing when exposed to adversarial perturbations of inputs \cite[e.g,][]{Goodfellow2015ExplainingAH,Szegedy2014IntriguingPO,Jia2017AdversarialEF,glockner-etal-2018-breaking,Dinan2019BuildIB}. 
Due to the opaque nature of neural modeling, methods for adversarial example generation may also steer algorithms towards generating unlikely examples that exhibit unrealistic properties \cite{Zhao2018GeneratingNA}.

In this work, we pose the question, \textit{``what does it really mean for an adversarial attack to be effective and can naturalistic adversaries match artificial ones?"} We argue that effectiveness should be dependent not only on attacking accuracy, but on usefulness of adversaries for improving robustness under realistic conditions 
\cite[e.g identifying social biases learned by neural models,][]{Buolamwini2018GenderSI,Sheng2019TheWW,stanovsky-etal-2019-evaluating,Sap2019TheRO,ross-etal-2021-measuring}. 

\begin{figure}[ht!]
    \centering
    \includegraphics[width=\linewidth]{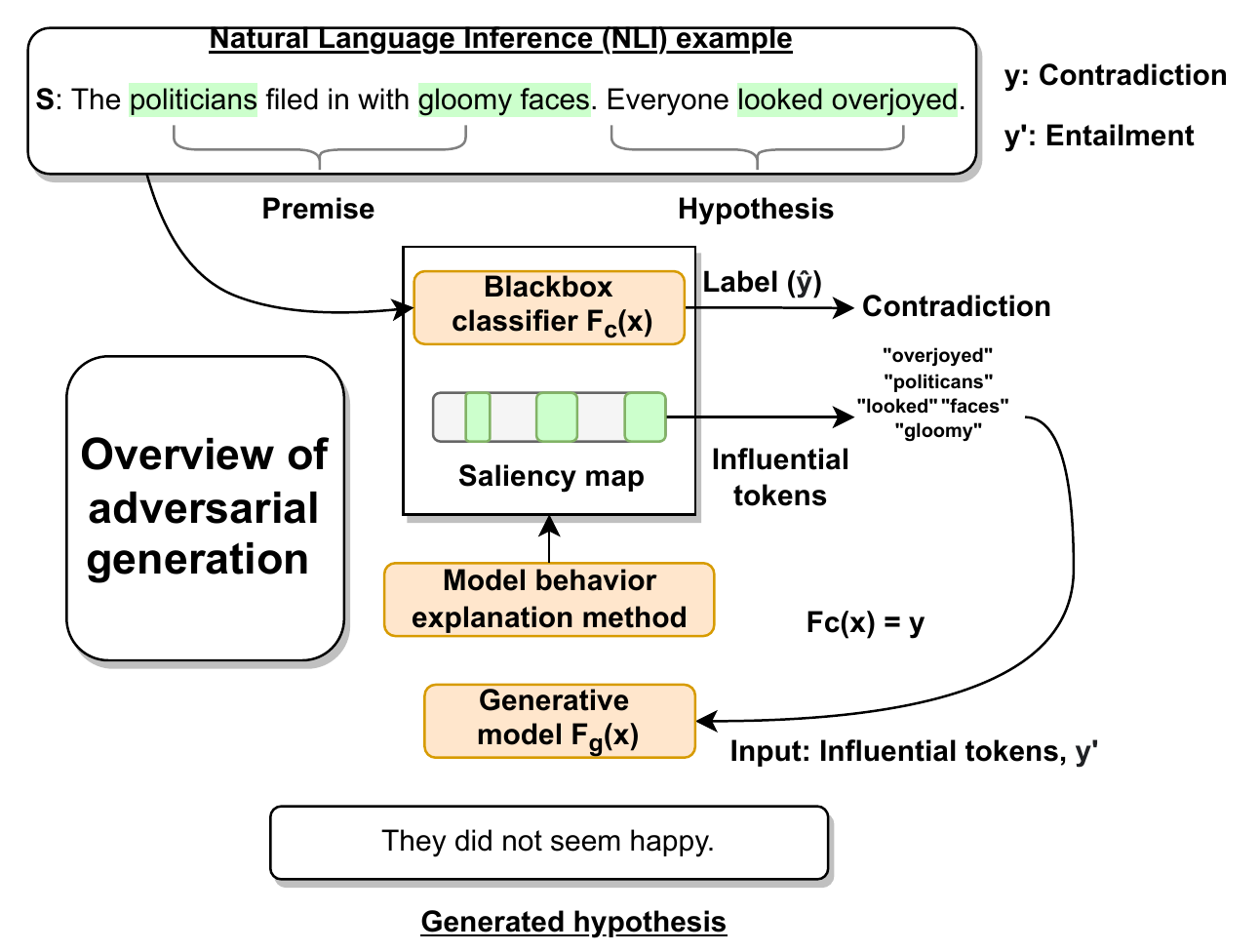}
    \caption{Our proposed framework for testing robustness of models using machine-generated adversarial examples (NaturalAdversaries). In the initial step, we probe the behavior of a black-box classifier (e.g. RoBERTa) using an explainability method like integrated gradients to find tokens with high attribution weights (influential tokens). We then use a generative model (e.g. GPT-2) to produce new adversarial examples conditioned on these tokens. 
    }
    \vspace{-4 mm}
    \label{fig:intro-fig}
\end{figure}

We propose a framework NaturalAdversaries\footnote{Code and data can be found here: \url{https://github.com/skgabriel/NaturalAdversaries}.} for generating convincingly naturalistic adversaries. We first approximate the behavior of a given classifier decision function $F_{c}(x)$ and then train a generative model $F_{g}(x)$ to mimic this behavior. As shown in Figure \ref{fig:intro-fig}, we condition generative models on influential tokens extracted using black box or white box explainability methods \cite{Ribeiro2016WhySI,Sundararajan2017AxiomaticAF}
, and a desired label (e.g. “\textit{entailment}'' or “\textit{contradiction}''), to produce new examples that match a distribution learned from $F_{c}(x)$ through sampling.



\begin{figure*}
\centering
\begin{subfigure}{.5\textwidth}
  \centering
  \includegraphics[width=1\linewidth]{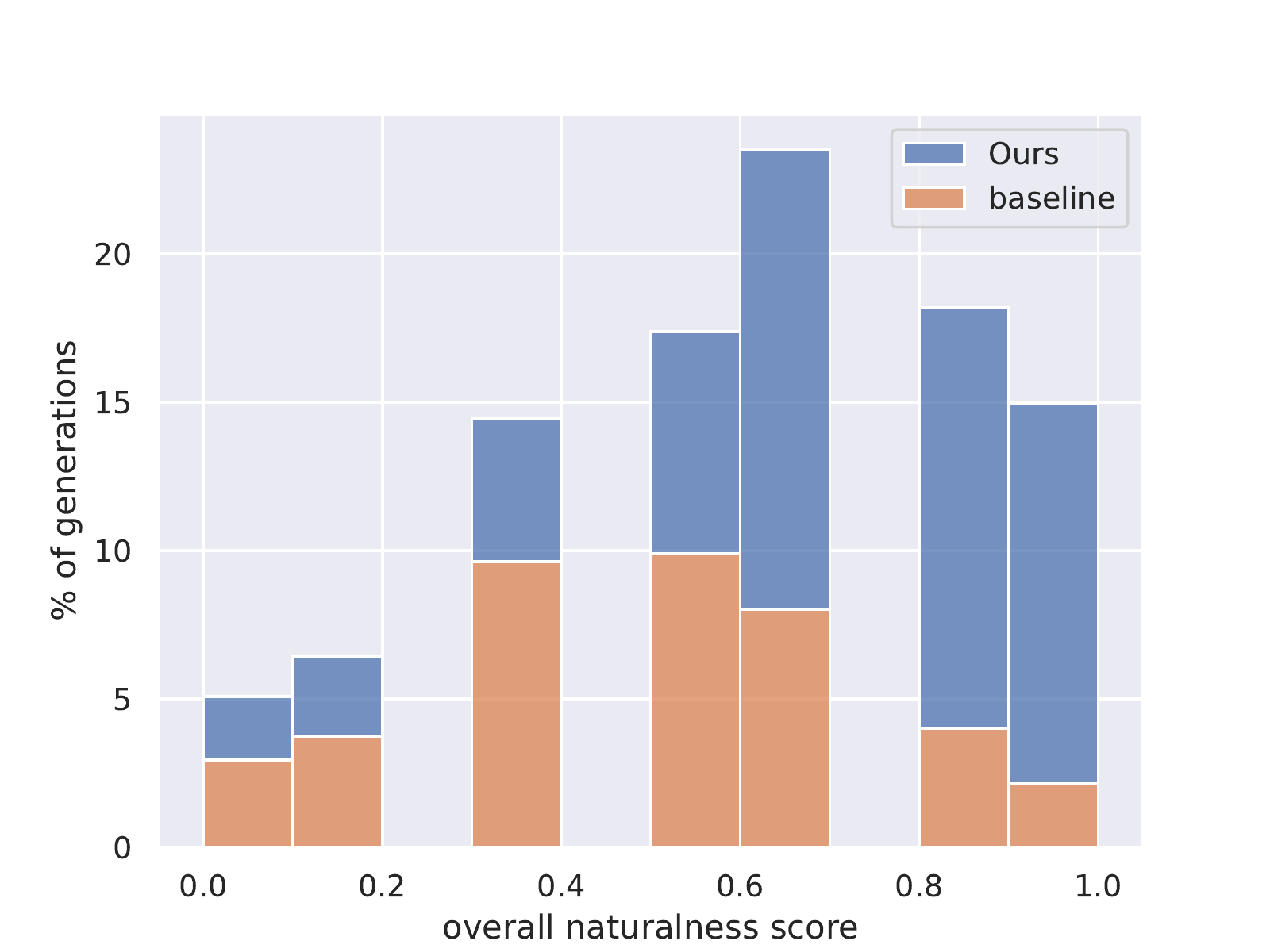}
  \caption{Distribution for ineffective examples.}
  \label{fig:sub1}
\end{subfigure}%
\begin{subfigure}{.5\textwidth}
  \centering
  \includegraphics[width=1\linewidth]{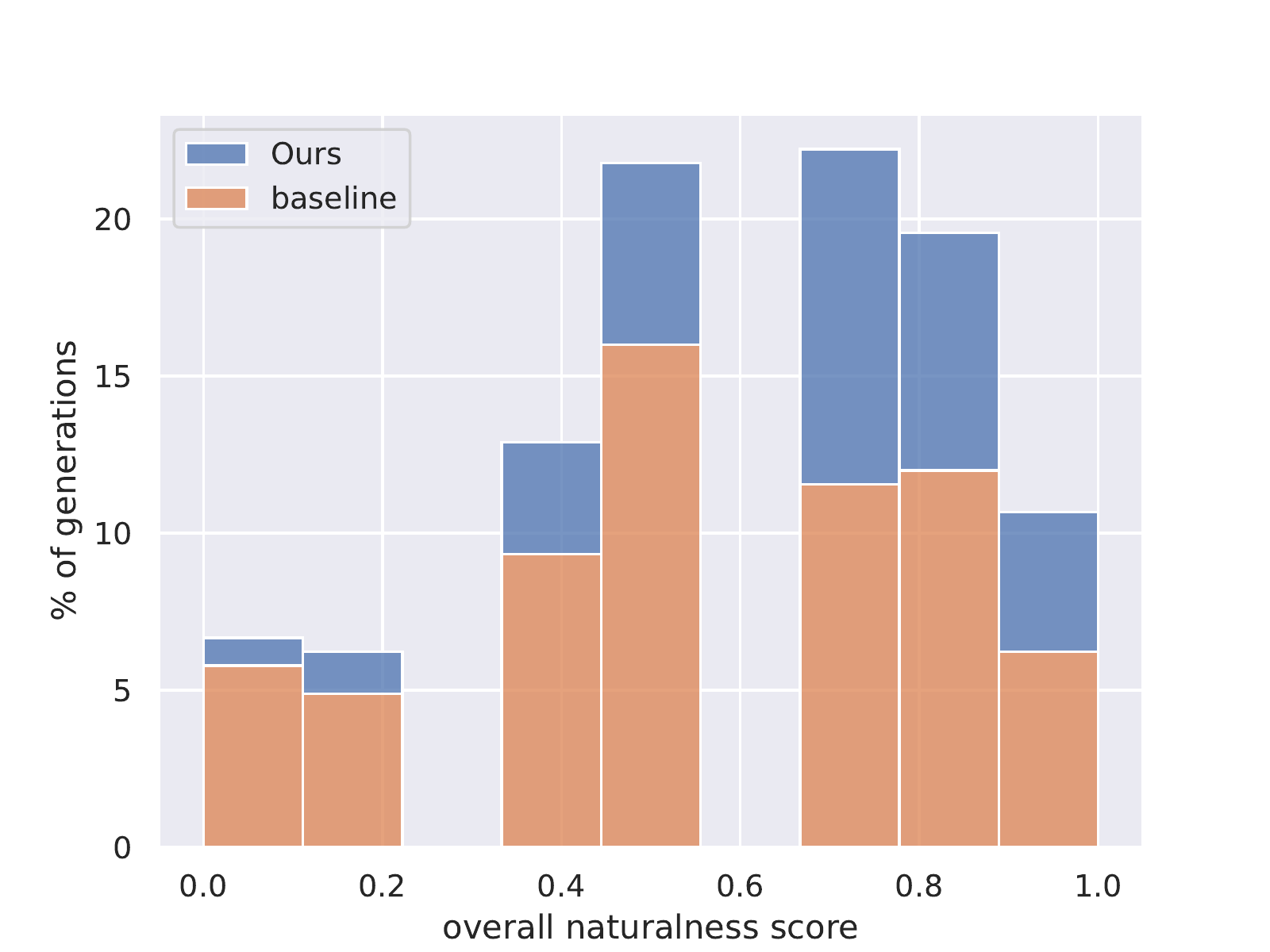}
  \caption{Distribution for effective examples.}
  \label{fig:sub2}
\end{subfigure}
\caption{Averaged naturalness scores from human evaluation. The adversarial examples were generated from the DynaHate test set \cite{vidgen-etal-2021-learning} using two common baselines (baseline) \cite{Ebrahimi2018HotFlipWA,Jin2020IsBR} as well as NaturalAdversaries (Ours). We show the distribution of scores for examples that are effective or ineffective respectively at fooling a RoBERTa toxicity classification model \cite{zhou-etal-2020-debiasing}. This shows that not only does NaturalAdversaries generate more natural examples, but naturalistic examples can also be effective at demonstrating adversarial behavior.} 
\label{fig:test}
\end{figure*}

Our results on two different tasks (\textit{hate speech detection} and  \textit{natural language inference}) show that our approach leads to adversaries that are perceived by annotators as considerably more natural. While naturalistic adversaries are often less adversarial than artificial adversaries \cite[following prior literature, e.g.][]{morris-etal-2020-reevaluating}, we find this depends on the evaluation setting and they can be better defenses.

\section{Defining Naturalness}

We first define what it means for machine-generated adversaries to have the quality of ``naturalness.'' Prior work on text generation has defined this property in terms of linguistic competence \cite{novikova-etal-2018-rankme,10.1162/tacl_a_00315}, as well as enumerating undesirable characteristics that lower perceived naturalness like self-contradiction \cite{Dou2022IsGT}. In our work, we ask human evaluators to judge naturalness in terms of whether generated text fragments are \textit{coherent}, \textit{well-formed} and \textit{likely to be human-written}.   

\section{Description of NaturalAdversaries}

Our overall framework consists of two stages - (1) a probing stage where we identify the influential (high attribution) tokens and (2) an adversarial generative stage where we generate unseen challenging examples by conditioning on the extracted tokens and a reversed label (\S\ref{sec:adgen}). 
We focus on two types of explainability methods as a means of summarizing model behavior through sampling - (1) LIME local linear explanation models \cite{Ribeiro2016WhySI} and (2) gradient attribution scores \cite{Sundararajan2017AxiomaticAF}. Using these methods as part of our proposed method is advantageous since it doesn't require curation of cherry-picked examples to probe model behavior, and is agnostic to the specific internal structure of a given classifier. Given a sequence of text tokens $S=[s_0, s_1, \dots, s_i, s_{i+1}, \dots]$ with classifier label $\hat{y}$, each of these methods define a scoring function $F_{attr}(s_i)$ which we use for assigning attribution scores to each token $s_i$ which represents its overall contribution to the classifier's decision. We separate these approaches based on whether $F_{attr}$ is conditioned on the model parameters or not. If it is, this demonstrates a white-box attack that can adapt to the vulnerabilities of a specific classifier given complete access to the model (e.g., using gradient attribution scores, see \S\ref{sec:white-box} in the Appendix for details). In the black-box setting (e.g. using LIME attribution scores), the underlying architecture and parameters are not known, and only sampled predictions are used to approximate the model behavior (see \S\ref{sec:black-box} in the Appendix). 

\begin{table*}[t]
\begin{center}
\begin{tabular}{p{5cm}p{3cm}p{4.25cm}p{1cm}p{1cm}}
Dataset & Taxonomy & Example & Classifier(s)\\ 
\midrule
DynaHate \cite{vidgen-etal-2021-learning}  & hate / nothate & I say I like women, but I don't & RoBERTa, BERT \\
ANLI \cite{nie-etal-2020-adversarial}  & contradiction / neutral / entailment & \textbf{P}: P-17 is a mixed use skyscraper proposed for construction in Dubai... The design is for a 379 m tall building, comprising 78 floors. \textbf{H}: P-17 is designed to have 78 floors and be over 500 meters tall. & DeBERTa, BERT\\

\end{tabular}
\caption{Description of considered datasets. For ANLI, each example comes with a premise (\textbf{P}) providing context and a hypothesis statement (\textbf{H}).}
\label{tab:domains}
\end{center}
\end{table*}

\subsection{Domains}
\label{sec:data}

\subsection{Adversarial Generation}
\label{sec:adgen}


Given a generative autoregressive model $F_g$ and training set $D_1$ with triples $(S,y,F_{attr}(S))$, we construct the following input sequence 
\begin{align}
x = [attr,z,label,y',text,S,eos]
\label{eq:2}
\end{align}
where $z$ is a sequence of influential tokens sampled from $S$ using the attribution weights defined by $F_{attr}(S)$, $attr$ is a special token indicating the start of this sequence, $y'$ is the desired classification label, $label$ and $text$ are special tokens indicating the start of $y'$ and $S$ respectively, and $eos$ is a special token indicating where the full input sequence ends. At training time, $y'=y$ as the generated model is trained to mimic the behavior of the classifier model and generate examples with a given label $y'$ based on the classifier's observed behavior. At decoding time, we encourage adversarial behavior by reversing the label (e.g. setting $y' = 1-y$). The model is prompted using only the influential tokens $z$ and $y'$. For example, given a natural language inference (NLI) premise and hypothesis pair (\textit{``It was sunny outside''}, \textit{``it was too dark to see anything outside''}) where the gold label is contradiction, at training time we use ($y'$="contradiction";$z$=``influentialWord1'', ``influentialWord2'', ``influentialWord3'', $S$=\textit{``It was sunny outside. It was too dark to see anything outside.''}). At decoding time we would use ($y'$="entailment";$z$=``influentialWord1'', ``influentialWord2'', ``influentialWord3'') and predict $S$.

We minimize cross-entropy loss during training time: 
\begin{align}
\mathcal{L}_{CE} = - \frac{1}{|S|} \sum_{i=1}^{|S|} log P(S_i|S_1,...,S_{i-1}). \label{eq:3}   
\end{align}

\section{Experimental Setup}
In this section we first introduce the domains we test on  (\S\ref{sec:data}) and then methods used for baseline comparison (\S\ref{sec:base_models}). All adversarial generators are based on the GPT-2 124M parameter model. We describe evaluation setups (\S\ref{sec:human}), as well as out-of-distribution evaluation (\S\ref{sec:ood}).

An advantage of the proposed generative method is that we can automatically extend human-in-the-loop adversarial generation methods like Adversarial NLI \cite[ANLI,][]{nie-etal-2020-adversarial}, which are costly and time-consuming to curate. Given this motivation, we focus on particularly challenging human-in-the-loop examples rather than cases that are already well solved by existing benchmarks. To study effectiveness of the proposed approach, we conducted experiments on the hate speech detection (DynaHate \cite{Vidgen2021LearningFT}) and natural language inference (NLI). For DynaHate we use a RoBERTa classifier trained on tweets \cite{Founta_Djouvas_Chatzakou_Leontiadis_Blackburn_Stringhini_Vakali_Sirivianos_Kourtellis_2018,zhou-etal-2020-debiasing}.
We test generalization across both model architectures and (non-adversarial) data domains using a BERT model \cite{devlin-etal-2019-bert} trained on the HateXplain dataset \cite{Mathew2021HateXplainAB}. 
For NLI we use DeBERTa \cite{He2021DeBERTaDB} trained on MNLI \cite{N18-1101}. We test generalization using BERT trained on the QNLI dataset \cite{Wang2018GLUEAM}. Further details are provided in Table \ref{tab:domains} and Appendix \ref{sec:domains}.

\begin{table*}
\begin{center}
\begin{tabular}{clccccc} \toprule
Dataset & Model & $\text{Natural}_H$ (\%)  & Adv1 (\%) & Adv2 (\%) & HateCheck (F1)\\ \midrule
& Original  & -  & - & -  & 55.01\\
& TF  & 53 & \textbf{69}  & \textbf{59}  & 55.59\\
DynaHate & HF  & 27 & \underline{30} & \underline{55}  & \underline{55.92}  \\
& NA-LIME  & \underline{67} & \underline{30} & \underline{55} & 55.90 \\
& NA-IG  & \textbf{73} & 21 & 36 & \textbf{56.69}\\ \midrule
Dataset & Model  & $\text{Natural}_H$ (\%) & Adv1 (\%) & Adv2 (\%)  & SNLI-Hard (F1) \\ \midrule
& Original  & - & - & - & 76.95\\
& TF  & 57 & \textbf{57} & \textbf{46} & 76.82 \\
ANLI & HF  & 64 & \underline{33} & 38 & \textbf{76.98}\\
& NA-LIME  & \underline{73} & 31 & \underline{43} & \textbf{76.98}\\
& NA-IG  & \textbf{89} & 27 & 42 & \underline{76.97}\\
\bottomrule
\end{tabular}
\caption{Human evaluation ($\text{Natural}_H$) of naturalness, along with adversarial performance against the original target classifier $Adv1$ and an unseen classifier $Adv2$. In the last topright column we show macro-averaged F1 performance on HateCheck \cite{rottger-etal-2021-hatecheck} after finetuning RoBERTa on 150 adversarial examples, compared to the original performance. We conduct a similar experiment for NLI using the SNLI-Hard evaluation set \cite{Gururangan2018AnnotationAI} with results in the last bottomright column. We bold the best-performing model and underline the second best model.}
\label{tab:quality}
\end{center}
\end{table*}

\subsection{Baselines}
\label{sec:base_models}

For automatic adversarial example construction, we compare against several common adversarial example generation approaches which are designed for either \textbf{black-box} (model-agnostic) or \textbf{white-box} (model-dependent) attacks. Baselines are implemented using TextAttack \cite{Morris2020TextAttackAF}. 

\paragraph{Black-Box Baselines} We use the TextFooler \cite{Jin2020IsBR} algorithm for generating coherent adversaries by replacing high-importance words in original examples with words that preserve semantic similarity. 

\paragraph{White-Box Baselines} Since our approach has an advantage over other baselines in the white-box setting of utilizing knowledge about model parameters, we also compare against a word-level version of the widely used HotFlip gradient-based approach \cite{Ebrahimi2018HotFlipWA}.


\subsection{Evaluation Metrics}



\subsubsection{Human Evaluation}
\label{sec:human}

To compare effectiveness of automatic methods, adversarial examples are manually validated to determine the true label. We also assess \textit{naturalness} of examples, i.e. whether they are perceived as realistic examples that could be written by humans. 
We use 156 crowd-source workers from Amazon Mechanical Turk (MTurk) with prior experience validating hate speech \cite{sap2020socialbiasframes} and 79 workers with experience validating NLI data \cite{liu2022wanli}. We sample 150 examples using each approach (some approaches may impose constraints that are unsatisfied by all candidate transformed sentences, we also filter out examples that are already adversarial to avoid conflating adversarialness of original examples with effects of the transformation). Each example is judged by 3 different workers. For hate speech, we classify an example as toxic if at least one annotator considers it so. We achieve moderate inter-annotator agreement of Fleiss' $\kappa=.51$ for hate speech and $\kappa=.52$ for NLI.

\subsubsection{Out-of-Distribution Performance}
\label{sec:ood}

Here we frame domain adaptation as a few-shot learning problem, where the adversarial evaluation set represents training examples from outside the seen domain of the classifier. To test out-of-distribution (OOD) performance on hate speech data, we use the HateCheck test suite \cite{rottger-etal-2021-hatecheck}, which consists of test cases for 29 model functionalities relating to real-world concerns of stakeholders. For NLI we check OOD performance on the SNLI-Hard dataset \cite{Gururangan2018AnnotationAI}, which assesses common model vulnerabilities.  



\section{Results}


We discuss results for the TextFooler (TF) and HotFlip (HF) baselines along with our two model variations (NA-LIME and NA-IG). 

\paragraph{Quality and effectiveness of adversarial generations.} From Table \ref{tab:quality}, we can see that examples generated using NaturalAdversaries are perceived as more natural across domains (20\% more for hate speech and 25\% more for NLI). While attacking accuracy is generally lower than artificial adversaries, we also find that our black-box approach generalizes well to classifiers other than the original target model, sometimes matching or beating the performance of artificial baselines (notably, NA-LIME does 5\% better on NLI for $Adv2$ than HotFlip). 



\paragraph{OOD performance.}

Although NLI model performance is relatively unaffected by finetuning, when we assess the RoBERTa classifier using HateCheck, we find that the target model exhibits concerning vulnerabilities. Finetuning generally improves performance, though our NA-IG model leads to the most improvement (1.68 F1 over base performance). Given the small size of our evaluation set (150 examples), this indicates naturalistic adversarial examples may address classifier weaknesses, with minimal need for manual annotation. We provide examples of generations in Appendix \ref{sec:expgen}.

\section{Related Work}

\paragraph{Adversarial Attacks} Prior work on adversarial attacks focus primarily on time-consuming and costly manual annotation \cite{kiela-etal-2021-dynabench}, or automatic example construction that relies upon a predefined type of attack \cite[e.g. testing robustness to syntactic and lexical errors,][]{Belinkov2018SyntheticAN,mccoy-etal-2019-right,Gabriel2021GOFA,Wu2021PolyjuiceGC}. The effect of complex adversarial attacks in the hate speech domain is also relatively unexplored. While \citet{Rusert2022OnTR} recently address this, they do not consider naturalness.

\section{Conclusion}

We introduce a framework for generation of naturalistic adversaries that is effective for multiple neural classifiers and across domains. We encourage further work on how naturalistic adversaries may improve robustness in real-world settings.





\section{Limitations}
\label{sec:limit}

While it is well-known that transformer-based language models suffer from lexical biases \cite{Gururangan2018AnnotationAI}, it may be an oversimplification to say that a single keyword is independently the cause of a particular classification decision. It has been shown that language models may consider compositionality to some degree \cite{shwartz-dagan-2019-still,Baroni2019LinguisticGA}, and future work may explore explainability methods that take this into consideration \cite[e.g.,][]{ye-etal-2021-connecting}. Another limitation is that generative approaches are highly dependent on the decoding method of choice \cite{Holtzman2020TheCC}, and while this provides us more flexibility, it also leads to more variability in performance. 

\section{Ethics \& Broader Impact Statement}

\paragraph{General Statement} While there is a risk of any technologies aimed at mimicking natural language being used for malicious purposes, our work has wide-ranging potential societal benefit by improving fairness and real-world robustness of neural classifiers. Increasingly it has become clear that pretrained neural language models do not operate from a neutral perspective, and implicitly learn behaviors that pose real harm to users from training data \cite{Jernite2022DataGI}. We demonstrate that our framework is effective at generating adversaries that uncover model vulnerabilities for two well-studied domains (hate speech and NLI), and it is hypothetically extensible to other domains like automated fact-checking. Given the sensitive nature of toxic language and hate speech detection in particular, we strongly emphasize that the work is intended only for research purposes or improving robustness of automated systems. For data and code release, we include detailed model and data cards \cite{bender-friedman-2018-data,10.1145/3287560.3287596}. 

\paragraph{Annotation} Based on time estimates, the annotator wage is approx. \$10-\$16 per hour. All annotators were required to click a consent button before working on the tasks. For hate speech annotation, annotators were cautioned about the possibly disturbing nature of the content before being shown examples. We also provided crisis hot-line information in case of emotional distress. 

\section{Acknowledgements}

We thank Microsoft Research for supporting human evaluations. We also thank the anonymous reviewers for their helpful comments, as well as members of UW NLP and Greg Durrett for insightful discussions.

\clearpage
\bibliographystyle{acl_bib}
\bibliography{tex/acl2020.bib}

\clearpage
\appendix

\section{Explainability Method Details}
\subsection{Black-Box Setting}
\label{sec:black-box}


Given a classifier $F_c(x)$ and a set of initial seed examples $D$ with ground truth labels $y$, we predict classifier labels $\hat{y}$ and measure the contribution of each token $s_i$ in a given sequence $S \in D$ to classifier's prediction using LIME \cite{Ribeiro2016WhySI}. The LIME algorithm defines a local neighborhood around a point x representing S, $N(x)$, using a proximity measure $\pi_x$, and optimizes linear models $g \in G$ to jointly minimize the distance of decision functions $g$ and $F_c$ for $\tilde{x} \in N(x)$ and also the complexity of $g$. In this case: 

\begin{equation} 
    \mathcal{L}(F_c, g, \pi_x) = \sum_{\tilde{x},\tilde{x}' \in N(x)}\pi_x(\tilde{x})(f(\tilde{x})-g(\tilde{x}'))^2
\end{equation}
\begin{equation}
    F_{attr}(S) = \text{argmin}_{g \in G} \mathcal{L}(F_c,g,\pi_x) + \Omega(g)
\end{equation}
where $\mathcal{L}(F_c, g, \pi_x)$ is the locality-aware loss, $\Omega(g)$ is model complexity, and $\pi_x$ is an exponential kernel defined using cosine distance\footnote{S is embedded using a simplified one-hot feature vector for g(x). In perturbed examples randomly selected tokens are masked.}. We also tested Shapley additive explanation values \cite{NIPS20178a20a862} in early experimentation, but found that the results were less promising than LIME.

\subsection{White-Box Setting}
\label{sec:white-box}

Given a classifier $F_c(x)$ and a set of initial seed examples $D$ with labels $y$, we predict classifier labels $\hat{y}$ and measure the contribution of each token $S_i$ in an example sequence $S \in D$ to this final output decision using the following computation:
\begin{align}
& F_{attr}(S_i) = (x_i - x'_i) \times \int_{\alpha=0}^{1}f(\alpha)d\alpha \nonumber\\
&f(\alpha)=\frac{\partial F_c(x' + \alpha \times (x - x'))}{\partial x_i}.
\label{eq:1}
\end{align}
Here $x_i$ is the embedding of $S_i$. Following \cite{mudrakarta-etal-2018-model}, the baseline input embedding ($x'$) is defined by a sequence of pad tokens that is the same length as the input, since the embedded pad tokens should not be informative. $\frac{\partial F(x)}{\partial x_i}$ is the gradient of $F_c(x)$ with respect to $x_i$ \cite{Sundararajan2017AxiomaticAF}. After identifying contribution of each token, we can partition $D$ into two sets $D_1$ and $D_2$ based on the model behavior, where $D_1$ forms a subset representing the space of correctly predicted examples and $D_2$ consists of incorrect predictions. We use the examples and attribution weights from $D_1$ as training data for a generative model $F_g(x)$.

\section{Domains}
\label{sec:domains}

\subsection{Hate Speech Detection}

For hate speech detection, we train the AdversarialGen generation model on the DynaHate benchmark. \cite{Vidgen2021LearningFT}. We use a RoBERTa classifier trained on the twitter hate speech dataset \cite{Founta_Djouvas_Chatzakou_Leontiadis_Blackburn_Stringhini_Vakali_Sirivianos_Kourtellis_2018,zhou-etal-2020-debiasing}. DynaHate benchmark was chosen for training in our experiments since it was constructed using a human-and-machine-in-the-loop setup designed to reduce dataset biases and improve model generalizability. It also includes examples of implicit hate, which rely less on lexical cues. For this task, the input to the classifier model is a text document like the one shown below
\begin{center}
    x = [CLS] \textit{all I want is to not be treated like a second-class citizen} [SEP].
\end{center}
Here $[CLS]$ and $[SEP]$ denote classifier-specific special tokens. The output is a binary label (benign or harmful). 

\subsection{Natural Language Inference}

For natural language inference, we train the generation model on the ANLI dataset \cite{nie-etal-2020-adversarial}, which was constructed similarly to DynaHate with an iterative human-and-machine-in-the-loop process. We test a DeBERTa-base classifier \cite{He2021DeBERTaDB} trained on the Multi-Genre NLI (MNLI) corpus \cite{N18-1101}, which is a featured task in the CLUES and GLUE evaluation benchmarks \cite{Mukherjee2021CLUESFL,Wang2018GLUEAM}. The MNLI corpus consists of $\sim$433k diverse sentence pairs, however recent work has shown that MNLI-trained models are highly susceptible to adversarial attacks from crowd-source workers \cite{nie-etal-2020-adversarial}. For this task, the input to the classifier model is a premise sentence $s_p$ and hypothesis sentence $s_h$ like the ones shown below
\begin{align*}
    \text{x = [CLS]} \underbrace{\textit{she walks behind me}}_{s_p}\text{[SEP]}\\\underbrace{\textit{she walks in front of me}}_{s_h}\text{ [SEP]}.
\end{align*}
The output is a label specifying whether the premise \textit{contradicts} the hypothesis, \textit{entails} the hypothesis or is \textit{neutral} (the hypothesis could either be true or false given the premise). 



\section{Additional Implementation Details}
\label{sec:additional_details}

For all explainability methods we took the top 20\% of tokens with the highest attribution scores. For LIME, we set the maximum number of features to 20 and generate 2,000 samples for training. 

All models are trained on a single Quadro RTX 8000 GPU. Training time averages 1 hour per epoch with a batch size of 32 for NLI data and 1 minute per epoch with a batch size of 16 for hate speech data. Inference time is approx. 10 minutes. We use the 124M parameter GPT-2 model for all generators. Generations are sampled after one training and evaluation run. We used a learning rate of 2e-4 for all models, determined by a hyperparameter search tuned on the dev. loss with the DynaHate trained model and potential learning rates in the range of $lr\in[.2,\text{2e-6}]$. For the hate speech domain we use top-$k$ decoding with $k=10$, while for natural language inference (NLI) we use beam search decoding with a beam size of 10 (we tested $k$ and beam size values from 3-10). We also use a max length context size of 150 tokens for hate speech and 500 for NLI. For both domains, we use a learning rate of $lr=.0004$ and train until early stopping on the dev. set. 

We train the hate speech domain model using round 1 of DynaHate, and combine test sets from all rounds. We use round 3 for the NLI model and sample from all rounds for the test set. 

\subsection{Data Processing}

For human evaluation, we remove non-ASCII characters and special tokens like "<|endoftext|>." We also noticed that the HotFlip baseline tends to generate extra spaces, and removed unncessary spaces. Newline characters were replaced by "<linebreak>." All the data is in English.

\subsection{Annotator Instructions}
\paragraph{Hate speech}
\textit{Read a text fragment (potentially generated by an AI system) and tell us whether it is potentially toxic/harmful to any reader. Note that there is no right or wrong answer, and we are interested in your opinion. We use a broad definition of harmful speech, including but not limited to use of inappropriate language, insults, threats, stereotyping, hate speech, microaggressions, derogatory speech and obscene content.}

\textit{Please look carefully at the text fragment and consider whether or not it is harmful. During each HIT you will be asked to answer several questions, listed and explained below.}

\textit{ 1) Is the text fragment harmful/toxic? Our purpose is to understand how harm/toxicity can show up in language, we are not making statements about the actual content of the posts. Try to answer this question while considering a broad set of people from different backgrounds, not just your own.}

\textit{ 2) Does the text fragment read naturally (Is it coherent, well-formed and likely human-written)?}

\paragraph{NLI}
\textit{Read a premise text fragment and a hypothesis text fragment (potentially generated by an AI system) and tell us what type of relationship there is between the two text fragments (entailment, contradiction, neutral). We provide detailed definitions and examples for these three types of relationships below.}

\textit{If you do not understand the text fragments, please indicate so using the given checkbox, but still answer the questions to the best of your abilities.}

\textit{\textbf{Entailment}: The two text fragments have this type of relationship if the information in the hypothesis follows from the premise (and no information contradicts the premise), e.g. the hypothesis "the woman is playing basketball" is entailed by the premise "the woman and the man are playing basketball in the park."}

\textit{\textbf{Contradiction}: The two text fragments have this type of relationship if at least some information in the hypothesis is contradicted by the premise, e.g. "the woman is playing soccer" is contradicted by the premise "the woman and the man are playing basketball in the park."}

\textit{\textbf{Neutral}: The two text fragments have this type of relationship if the information in the hypothesis is neither entailed by or contradicted by the premise (the two text fragments may be completely unrelated), e.g. the relationship between "the woman and the man are playing basketball in the park" and "the woman loves basketball" is neutral.}

\textit{\textbf{Questions:}}

\textit{1) What is the relationship (entailment, contradiction or neutral) between the two text fragments?}

\textit{2) Do the premise and hypothesis text fragments read naturally (are they coherent, well-formed and likely human-written)?}

\begin{table*}
\begin{center}
\begin{tabular}{ |p{8cm}|c|c|} \toprule
\textbf{Example} & \textbf{Domain}  & \textbf{Gold Label} \\ \toprule
Kobo Town is a Canadian Caribbean music group, led by Trinidadian Canadian singer and songwriter Drew Gonsalves. Based in Toronto, Ontario... [SEP] \textbf{Kobo Town is based in Canada.} & \textbf{NLI} & 2 (Entailment)\\
 & &\\
Snake Handler<br>Clark went to church with his second cousin. Clark was shocked when the preacher held a poisonous snake...[SEP] \textbf{Clark is named after a preacher.}  & \textbf{NLI}  & 1 (Neutral) \\ 
 & &\\
"How to ask about someone\'s disability<br>Consider whether the information is relevant. If it\'s not really relevant to you, then asking is probably rude...[SEP] \textbf{People with disabilities are generally brave.} & \textbf{NLI}  & 1 (Neutral) \\ 
 & &\\
I\'m glad that the vaccine is so popular, but I wonder if this would be a good thing for our democracy. & \textbf{Hate Speech}  &  1 (Toxic/Hate)  \\ 
 & &\\
I've always hated the word "feminists" because they are basically feminists, but I hate them for their views on women. & \textbf{Hate Speech}  &  1 (Toxic/Hate)  \\ 
 \bottomrule
\end{tabular}
\caption{Examples of NaturalAdversaries generations that were misclassified. For NLI only the hypothesis is generated, which we highlight in bold.} 
\label{tab:examples}
\end{center}
\end{table*}

\section{Generation Examples}
\label{sec:expgen}

We provide examples of generated examples in Table \ref{tab:examples}.

\end{document}